\documentclass[twoside,leqno,twocolumn]{article}  
\usepackage{ltexpprt} 

\usepackage{cite}
\usepackage{amsmath}
\usepackage{booktabs}
\usepackage[table]{xcolor}
\usepackage{graphicx}
\usepackage{subfigure}

\begin{document}

\title{Grounded Discovery of Coordinate Term Relationships\\between Software
Entities}

\author{Dana Movshovitz-Attias \\
Computer Science Department \\
Carnegie Mellon University \\
\tt{dma@cs.cmu.edu}
\\
\and
William W. Cohen \\
Machine Learning Department \\
Carnegie Mellon University \\
\tt{wcohen@cs.cmu.edu} }
\date{}

\maketitle

\begin{abstract}
We present an approach for the detection of coordinate-term relationships
between entities from the software domain, that refer to Java
classes.
Usually, relations are found by examining corpus statistics associated with text
entities. In some technical domains, however, we have access to additional
information about the real-world objects named by the entities, suggesting that
coupling information about the ``grounded'' entities with corpus statistics
might lead to improved methods for relation discovery. To this end, we develop a
similarity measure for Java classes using distributional information about how
they are used in software, which we combine with corpus statistics on the
distribution of contexts in which the classes appear in text.
Using our approach, cross-validation accuracy on this dataset can be
improved dramatically, from around 60\% to 88\%. Human labeling results show
that our classifier has an F1 score of 86\% over the top 1000 predicted pairs.
 
\end{abstract}

\section{Introduction}

Discovering semantic relations between text entities is a key task in natural
language understanding. 
It is a critical component which
enables the success of knowledge representation systems such as TextRunner
\cite{yates2007textrunner}, ReVerb \cite{fader2011identifying}, and NELL
\cite{carlson2010toward}, which in turn are useful for a
variety of NLP applications, including, temporal scoping
\cite{talukdar2012coupled}, semantic parsing \cite{krishnamurthy2012weakly} and entity linking
\cite{lin2012entity}.

In this work, we examine \emph{coordinate} relations between words. According to
the WordNet glossary, \emph{X} and \emph{Y} are defined as \emph{coordinate
terms} if they share a common hypernym \cite{miller1995wordnet,
christiane1998wordnet}. This is a symmetric relation that indicates a semantic
similarity, meaning that \emph{X} and \emph{Y} are ``a type of the same
thing'', since they share at least one common ancestor in some hypernym taxonomy (to
paraphrase the definition of Snow et al. \cite{snow2006semantic}).

Semantic similarity relations are normally discovered by comparing corpus
statistics associated with the entities: for instance, two entities $X$ and $Y$ that usually appear in similar
contexts are likely to be semantically similar
\cite{pereira1993distributional,pantel2003clustering,curran2004distributional}.
However, in technical domains, we have access to additional
information about the real-world objects that are named by the entities: e.g.,
we might have biographical data about a person entity, or a 3D structural encoding of a protein entity. In
such situations, it seems plausible that a "grounded" NLP method, in which
corpus statistics are coupled with data on the real-world referents of $X$ and
$Y$, might lead to improved methods for relation discovery.

Here we explore the idea of grounded relation discovery in the domain of
software. 
In particular, we consider the detection of coordinate-term
relationships between entities that (potentially) refer to Java classes.
We use a software domain text corpus derived from
the Q\&A website StackOverflow (SO), in which users ask and answer questions
about software development, and we extract posts which have been labeled by
users as \emph{Java} related. From this data, we collected a small set of
entity pairs that are labeled as coordinate terms (or not) based
on high-precision Hearst patterns and frequency statistics, and we attempt to
label these pairs using information available from higher-recall approaches based on
distributional similarity.

\begin{figure*}[t]
\centering
	\includegraphics[width=0.85\textwidth]{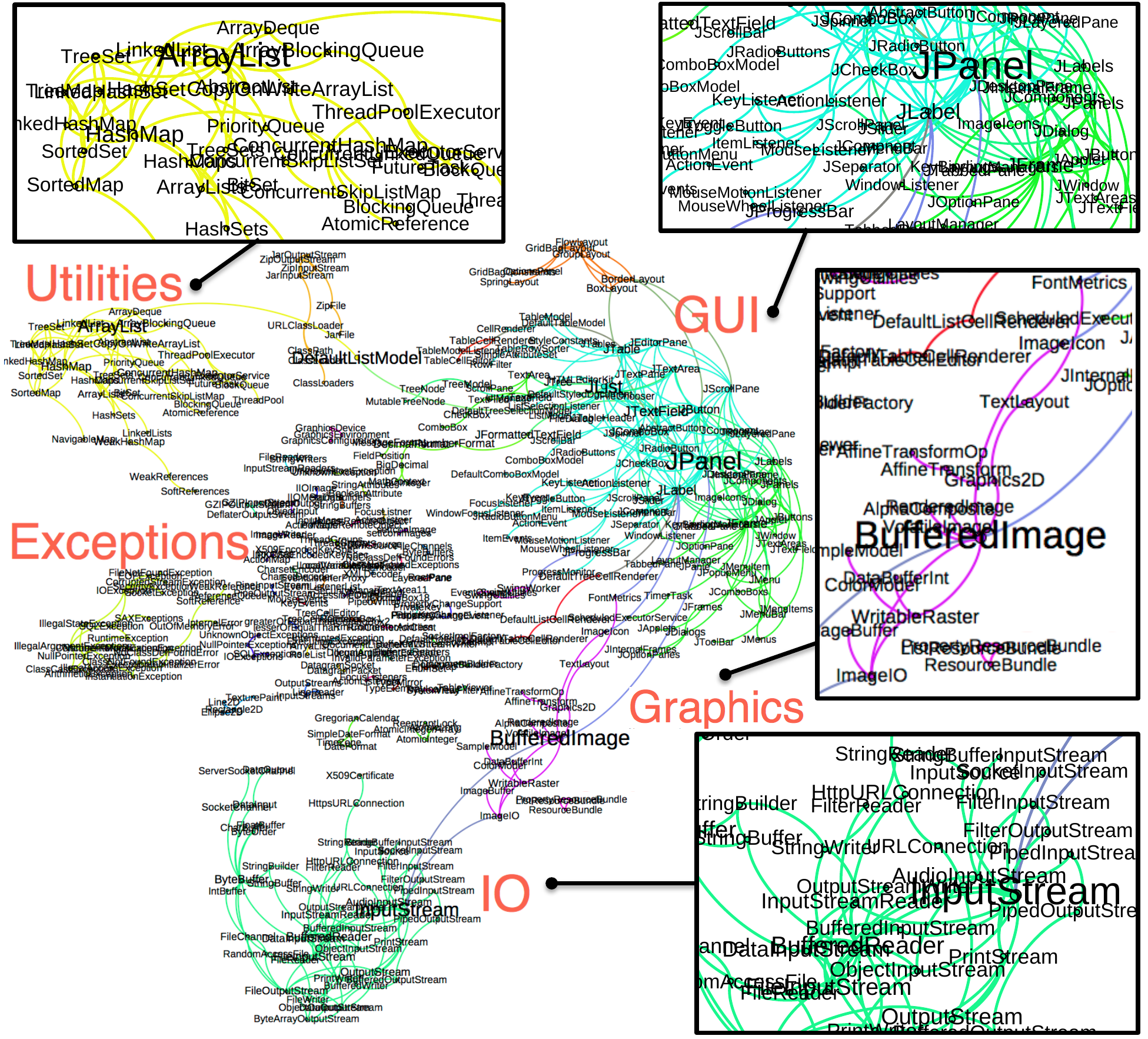}
	\caption{Visualization of predicted coordinate term pairs, where each pair of
	coordinate classes is connected by an edge. Highly connected components
	 are labeled by edge color, and it can be noted that they contain classes with
	 similar functionality. Some areas containing a functional class group have
	 been magnified for easier readability.}
  \label{fig:topology}
\end{figure*}

We describe an entity linking method in order to map a given text entity to an
underlying class type implementation from the Java standard libraries. Next, we
describe corpus and code based information that we use for the relation
discovery task. Corpus based methods include distributional similarity and
string matching similarity.
Additionally, we use two sources of code based information: (1) we define the
\emph{class-context} of a Java class in a given code repository, and are
therefore able to calculate a code-based distributional similarity measure for
classes, and (2) we consider the hierarchical organization of classes, described
by the Java class type and namespace hierarchies.
We demonstrate that using our approach, cross-validation accuracy on this
dataset is improved from 60.9\% to 88\%. According to human labeling, our
classifier has an F1-score of 86\% over the highest-ranking 1000 predicted
pairs.

We see this work as a first step towards building a knowledge
representation system for the software domain, in which text entities refer to
elements from a software code base, for example classes, methods,
applications and programming languages.
Understanding software entity relations will allow the construction of a domain
specific taxonomy and knowledge base, which can enable higher reasoning
capabilities in NLP applications for the software domain
\cite{weimer2007automatically,wang2009extracting,branavan2010reading,movshovitzattias-wcohen:2013:ACL} and improve a variety of code assisting applications, including code refactoring
and token completion
\cite{han2009code,jacob2010code,binkley2011improving,schulam:2013:DAPSE13p1}.

Figure~\ref{fig:topology} shows a visualization based on coordinate term
pairs predicted using the proposed method. Java
classes with similar functionality are highly connected in this graph, indicating that our
method can be used to construct a code taxonomy.

\section{Related Work}

\textbf{Semantic Relation Discovery.}
Previous work on semantic relation discovery, in particular, 
coordinate term discovery, has used two main
approaches. The first is based on the insight that certain lexical patterns
indicate a semantic relationship with high-precision, as initially observed by
Hearst \cite{hearst1992automatic}. For example, the conjuction pattern ``X and Y''
indicates that $X$ and $Y$ are coordinate terms. Other
pattern-based classifier have been introduced for meronyms
\cite{girju2003learning}, synonyms \cite{lin2003identifying}, and general
analogy relations \cite{turney2003combining}.
The second approach relies on the notion that words that appear in a similar
context are likely to be semantically similar. In contrast to pattern based
classifiers, context distributional similarity approaches are
normally higher in recall.
\cite{pereira1993distributional,pantel2003clustering,curran2004distributional,snow2004learning}.
In this work we attempt to label samples extracted with high-precision Hearst
patterns, using information from higher-recall methods.

\textbf{Grounded Language Learning.} The aim of grounded language learning
methods is to learn a mapping between natural language (words and sentences) and
the observed world
\cite{siskind1996computational,yu2004integration,gorniak2007situated}, where
more recent work includes grounding language to the physical world
\cite{krishnamurthy2013jointly}, and grounding of entire discourses \cite{minh2013parsing}.
Early work in this field relied on supervised aligned sentence-to-meaning data
\cite{zettlemoyer2005learning,ge2005statistical}. However, in later work the
supervision constraint has been gradually relaxed
\cite{kate2007learning,liang2009learning}. 
Relative to prior work on grounded language
acquisition, we use a very rich and complex representation of entities and
their relationships (through software code). However, we consider a very
constrained language task, namely coordinate term discovery.

\textbf{Statistical Language Models for Software.} 
In recent work by NLP and software engineering researchers, statistical language
models have been adapted for modeling software code. NLP models have been used
to enhance a variety of software development tasks such as code and comment
token completion
\cite{han2009code,jacob2010code,movshovitzattias-wcohen:2013:ACL,schulam:2013:DAPSE13p1},
analysis of code variable names \cite{lawrie2006s,binkley2011improving}, and
mining software repositories \cite{gabel2008javert}. This has been complemented
by work from the programming language research community for structured
prediction of code syntax trees \cite{omar2013structured}. To the best of our
knowledge, there is no prior work on discovering semantic relations for software
entities.

\section{Coordinate Term Discovery}

\begin{figure}[t]
\centering
	\includegraphics[width=0.5\textwidth]{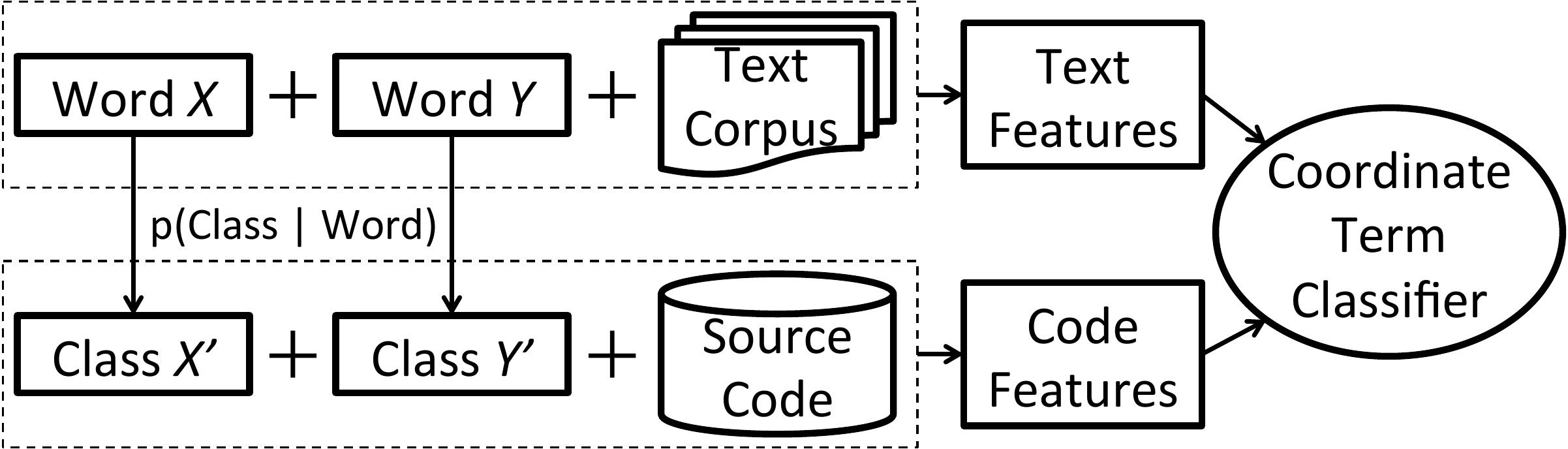}
	\caption{Classification Pipeline for determining whether
	nouns $X$ and $Y$ are coordinate terms. Each noun is mapped to an underlying 
	class from the code repository with probability,
	$p(\textrm{Class} | \textrm{Word})$. Textual features are extracted based on
	the input words, code based features are extracted using the mapped classes, and all of these are given to
	the coordinate term classifier. }
  \label{fig:classificationPippline}
\end{figure}

In this section we describe a coordinate term classification pipeline,
as depicted at high-level in Figure~\ref{fig:classificationPippline}. All the
following steps are described in detail in the sections below. 

Given a software domain text corpus (StackOverflow) and a code
repository (Java Standard Libraries), our goal is to predict a coordinate
relation for $\langle X,Y \rangle$, where $X$ and $Y$ are nouns which
potentially refer to Java classes.

We first attempt a baseline approach of labeling the pair $\langle X,Y \rangle$
based on corpus distributional similarity. Since closely related classes often
exhibit morphological closeness, we use as a second baseline the string
similarity of $X$ and $Y$.

Next, we map noun $X$ to an underlying class implementation from the code
repository, named $X'$, according to an estimated probability for
$p(\textrm{Class }X' | \textrm{Word }X)$, s.t., $X'=\max_C{\hat{p}(C|X)}$, for
all other classes $C$. $X'$ is then the code referent of $X$. Similarly, we map
$Y$ to the class $Y'$.
Given a code-based grounding for $X$ and $Y$ we extract information using the class implementations: (1) we
define a code based distributional similarity measure, using code-context to
encode the usage pattern of a class, and (2) we use the hierarchical
organization of classes, described by the type and namespace hierarchies.
Finally, we combine all the above information in a single SVM
classifier.

\subsection{Baseline: Corpus Distributional
Similarity.}\label{sec:text-dist-sim}

As an initial baseline we calculate the corpus distributional similarity of
nouns $\langle X,Y \rangle$, following the assumption that words with similar
context are likely to be semantically similar.
Our implementation follows Pereira et al. \cite{pereira1993distributional}.
We calculate the empirical context distribution for noun $X$
\begin{equation}\label{eq:empirical-context-dist}
p_X = f(c,X) / \sum_{c'} f(c',X)
\end{equation}
where $f(c,X)$ is the frequency of occurrence of noun $X$ in context $c$. We
then measure the similarity of nouns $X$ and $Y$ using the \emph{relative
entropy} or \emph{Kullback-Leibler divergence}
\begin{equation}
D(p_X || p_Y) = \sum_z p_X(z)\log\frac{p_X(z)}{p_Y(z)}
\end{equation}
As this measure is not symmetric we finally consider the distributional
similarity of $X$ and $Y$ as $D(p_X || p_Y) + D(p_Y || p_X)$.

\subsection{Baseline: String Similarity.}\label{sec:string-sim}

Due to naming convention standards, many related classes often exhibit some
morphological closeness. For example, classes that provide Input/Output access
to the file system will often contain the suffix \texttt{Stream} or
\texttt{Buffer}. Likewise, many classes extend on the names of their super
classes (e.g., \texttt{JRadioButtonMenuItem} extends the class
\texttt{JMenuItem}). More examples can be found in
Figure~\ref{fig:topology} and Table~\ref{tbl:pairs-top-predicted}. We therefore
include a second baseline which attempts to label the noun pair $\langle X,Y
\rangle$ as coordinate terms according to their string matching similarity.
We use the SecondString open
source Java toolkit\footnote{http://secondstring.sourceforge.net/}. Each string
is tokenized by camel case (such that \emph{ArrayList} is represented as
\emph{Array List}). We consider the SoftTFIDF distance of
the tokenized strings, as defined by Cohen et al. 
\cite{cohen2003comparison}.

\subsection{Entity Linking.}\label{sec:text-to-code-map}

In order to draw code based information on text entities, we
define a mapping function between words and class types. Our goal is to find
$p(C|W)$,
where $C$ is a specific class implementation and $W$ is a word. This mapping is
ambiguous, for example, since 
users are less likely to mention the qualified class name (e.g., \verb|java.lang.String|), and usually use
the class \emph{label}, meaning the name of the class not including its
package (e.g., \verb|String|). As an example, the terms
\verb|java.lang.String| and \verb|java.util.Vector| appears 37 and 1 times
respectively in our corpus, versus the terms \verb|String| and \verb|Vector|
which appear 35K and 1.6K times. Additionally, class names appear with
several variations, including, case-insensitive versions, spelling mistakes, or
informal names (e.g., \emph{array} instead of \emph{ArrayList}).

Therefore, in order to approximate $p(C,W)$ in 
\begin{equation}
	p(C|W) = \frac{p(C,W)}{p(W)}
\end{equation}
We estimate a word to class-type mapping that is
mediated through the class label, $L$, as
\begin{equation}
	\hat{p}(C,W) = p(C,L) \cdot p(L,W)
\end{equation}
Since $p(C,L) = p(C|L)p(L)$, this can be
estimated by the corresponding MLEs
\begin{eqnarray}
	\hat{p}(C,L) &=& \hat{p}(C|L)\cdot \hat{p}(L) \nonumber \\
	&=& \frac{f(C)}{\sum_{C'
	\in L}f(C')}\cdot
	\frac{f(L)}{\sum_{L'}f(L')}
\end{eqnarray}
where $f()$ is the frequency function. Note that since $\sum_{C' \in L}f(C') =
f(L)$ we get that $\hat{p}(C,L)=\hat{p}(C)$, as the class label is uniquely
determined by the class qualified name (the opposite does not hold since multiple class types
may correspond to the same label). Finally, the term $p(L,W)$ is estimated by
the symmetric string distance between the two strings, as described
in Section~\ref{sec:string-sim}. We consider the linking probability
of $\langle X,Y \rangle$ to be $\hat{p}(X'|X)\cdot \hat{p}(Y'|Y)$,
where $X'$ is the best matching class for $X$ s.t. $X'=\max_C{\hat{p}(C|X)}$ and similarly for
$Y'$.

\subsection{Code Distributional
Similarity.}\label{sec:software-dist-sim}

Corpus distributional similarity evaluates the occurrence of words in particular
semantic contexts. By defining the \emph{class-context} of a Java class, we can
then similarly calculate a \emph{code distributional similarity} between
classes. Our definition of class context is based on the usage of a class as an
argument to methods and on the API which the class provides, and it is detailed
in Table~\ref{tbl:code-context}. We observe over 23K unique contexts in our code
repository. Based on these definitions we can compute the distributional
similarity measure between classes $X'$ and $Y'$ based on their code-context distributions, as previously described for the corpus
distributional similarity (Section~\ref{sec:text-dist-sim}, following
Pereira et al. \cite{pereira1993distributional}).
For the code-based case, we calculate the empirical context distribution of $X'$
(see Equation~\ref{eq:empirical-context-dist}) using $f(c,X')$, the
occurrence frequency of class $X'$ in context $c$, where $c$ is one of
the ARG-\emph{Method} or API-\emph{Method} contexts (defined in
Table~\ref{tbl:code-context}) for methods observed in the code repository.
The distributional
similarity of $\langle X',Y' \rangle$ is then taken, using the relative entropy,
as $D(p_{X'} || p_{Y'}) + D(p_{Y'} || p_{X'})$.

\begin{table} [t]
\centering
\begin{tabular}{ p{3in} }
	\toprule
\textbf{ARG-\emph{Method}:} \verb|Class| is being passed as an
argument to \verb|Method|. We count an occurrence of this context
once for the method definition\\
\verb|    Method(Class class, ...)|\\
as well as for each method invocation\\
\verb|    Method(class, ...)|\\
For example, given the statement\\
\verb|    str = toString(i);|\\
where $i$ is an Integer, we would count an occurrence for
this class in the context ARG-toString.
\tabularnewline
\cmidrule{1-1}
\textbf{API-\emph{Method}:} \verb|Class| provides the API method \verb|Method|.
We count an occurrence of this context once for the method definition, and for
every occurrence of the method invocation, e.g. \verb|class.Method(...)|.\\
For example, given the statement\\
\verb|    s = map.size();|\\
where $map$ is a HashMap, we would count an occurrence for
this class in the context API-size.
\tabularnewline
  	\bottomrule
\end{tabular}
\caption{Definition of two types of code-contexts for a class type,
\texttt{Class}, or an instantiation of that type (e.g., \texttt{class}).}
\label{tbl:code-context}
\end{table}

\subsection{Code Hierarchies and Organization.}\label{sec:code-hierarchies}

The words $X$ and $Y$ are defined as coordinate terms
if they have the same hypernym in a given taxonomy, meaning they have at least
one common ancestor in this taxonomy \cite{snow2004learning}.
For the purpose of comparing two class types, we therefore
define an ancestry relation between them using two taxonomies based on the
code namespace and type hierarchies. 

\textbf{Package Taxonomy:}
A package is the standard way for defining
namespaces in the Java language.
It is a mechanism for organizing sets of classes which normally share a common
functionality. Packages are organized in a hierarchical structure which can be
easily inferred from the class name. For example, the class
\verb|java.lang.String|, belongs to the
\verb|java.lang| package, which belongs to the \verb|java| package.

\textbf{Type Taxonomy:}
The inheritance structure of classes and interfaces in
the Java language defines a type hierarchy, such that class $A$ is the ancestor
of class $B$ if $B$ extends or implements $A$.

We define type-ancestry and package-ancestry relations between classes
$\langle X',Y' \rangle$, based on the above taxonomies. For the type
taxonomy,
\begin{quote}
$A_{type}^n(X',Y')$ = \{\# of common
ancestors $X'$ and $Y'$ share within \emph{n} higher up levels in the type
taxonomy\}
\end{quote}
for \emph{n} from 1 to 6. $A_{package}^n$ is defined similarly for the
package taxonomy. As an example,
\begin{equation*}
A_{package}^2(\textrm{ArrayList},
\textrm{Vector}) = 2 
\end{equation*}
as these classes both belong in the package \verb|java.util|, and therefore
their common level 2 ancestors are: \verb|java| and \verb|java.util|. Moreover, 
\begin{equation*}
A_{type}^1(\textrm{ArrayList},
\textrm{Vector}) = 5
\end{equation*}
since both classes extend the \verb|AbstractList| class, and also implement four
joint interfaces: \verb|List|, \verb|RandomAccess|,
\verb|Cloneable|, and \verb|Serializable|.

\section{Experimental Settings}

\subsection{Data Handling.}
We downloaded a dump of the interactions on the StackOverflow
website\footnote{http://www.clearbits.net/creators/146-stack-exchange-data-dump}
from its launch date in 2008 and until 2012.
We use only the 277K questions labeled with the user-assigned \emph{Java} tag,
and their 629K answers.

Text from the SO html posts was extracted with the Apache Tika
toolkit\footnote{http://tika.apache.org/} and then tokenized with the Mallet
statistical NLP package \cite{mccallum2002mallet}. In this study, we use only
the text portions of the SO posts, and exclude all raw code segments, as
indicated by the user-labeled $<$$code$$>$ markup. Next, the text was POS tagged
with the Stanford POS tagger \cite{toutanova2003feature} and parsed with the 
MaltParser \cite{nivre2006maltparser}. Finally, we extract noun pairs with the
conjunction dependencies: \emph{conj} or \emph{inv-conj}, a total of
255,150 pairs, which we use as positive training samples.

We use the Java standard libraries code repository as a grounding source for
Java classes, as we expect that users will often refer to these classes in
the \emph{Java} tagged SO posts. This data includes: 7072 source code files, the
implementation of 10562 class and interface types, and 477
packages. The code repository is parsed using the Eclipse JDT compiler tools,
which provide APIs for accessing and manipulating Abstract Syntax Trees.

\subsection{Classification.}\label{sec:coord-term-class}

We follow the classification pipeline described in
Figure~\ref{fig:classificationPippline}, using the LibLinear SVM classifier
\cite{fan2008liblinear,chang2011libsvm} with the following features:
\begin{description} 
\item[Corpus-Based Features] \hfill
	\begin{itemize} 
	\item \emph{Corpus distributional similarity} (Corpus Dist.
	Sim.) - see Section~\ref{sec:text-dist-sim}.
	\item \emph{String similarity} (String Sim.) - see
	Section~\ref{sec:string-sim}.
	\end{itemize}
\item[Code-Based Features] \hfill
	\begin{itemize} 
	\item \emph{Text to code linking probability} (Text-to-code
	Prob.) - see Section~\ref{sec:text-to-code-map}.
    \item \emph{Code distributional similarity} (Code Dist. Sim.) -
    see Section~\ref{sec:software-dist-sim}.
	\item \emph{Package and type ancestry} ($A_{package}^1$ -
	$A_{package}^6$ and $A_{type}^1$ - $A_{type}^6$) - see
	Section~\ref{sec:code-hierarchies}.
	\end{itemize}
\end{description}

\begin{table} [t]
\centering
\rowcolors{1}{white}{gray!13}
\begin{tabular}{ p{4.6cm} l }
	\toprule
	High PMI &
	Low PMI 
	\tabularnewline
	\cmidrule{1-1} \cmidrule(l){2-2}
  	\textbf{\small{$\langle$JTextField,JComboBox$\rangle$}} &
  	\small{$\langle$threads,characters$\rangle$}
  	\tabularnewline
  	\textbf{\small{$\langle$yearsPlayed,totalEarned$\rangle$}}&
  	\textbf{\small{$\langle$server,user$\rangle$}}
  	\tabularnewline
  	\textbf{\small{$\langle$PostInsertEventListener,} \newline
  	\small{PostUpdateEventListener$\rangle$}} &
  	\textbf{\small{$\langle$code,design$\rangle$}}
  	\tabularnewline
  	\textbf{\small{$\langle$removeListener,addListener$\rangle$}} &
  	\textbf{\small{$\langle$Java,client$\rangle$}}
  	\tabularnewline
  	\small{$\langle$MinTreeMap,MaxTreeMap$\rangle$} & \small{$\langle$Eclipse,array$\rangle$}
  	\tabularnewline
  	\bottomrule
\end{tabular} 
\caption{Sample set of word pairs with high and low \emph{PMI} scores. Many of
the high \emph{PMI} pairs refer to software entities such as variable,
method and Java class names, whereas
the low \emph{PMI} pairs contain more general software terms.}
\label{tbl:word-sample-pmi}
\end{table}

\begin{table} [t]
\centering
\begin{tabular}{ l c c }
	\toprule
	Method &
	\emph{Coord} &
	\emph{Coord-PMI}
	\tabularnewline
	\cmidrule{1-1} \cmidrule(l){2-2} \cmidrule(l){3-3}
	Code \& Corpus & \textbf{85.3} & \textbf{88}
  	\tabularnewline
  	\midrule
  	\emph{Baselines:}
  	\tabularnewline
  	\texttt{  }Corpus Dist. Sim. & 57.8 & 58.2
  	\tabularnewline
  	\texttt{  }String Sim. & 65.2 & 65.8
  	\tabularnewline
  	\texttt{  }Corpus Only & 64.7 & 60.9
  	\tabularnewline
  	\texttt{  }Code Only & 80.1 & 81.1
  	\tabularnewline
  	\midrule
  	\emph{Code Features:}
  	\tabularnewline
  	\texttt{  }Code Dist. Sim. & 67 (60.2) & 67.2 (59)
	\tabularnewline
	\texttt{  }$A_{package}^1$ & 64.2 (63.8) & 64.3 (63.9)
	\tabularnewline
	\texttt{  }$A_{package}^2$ & 64.2 (63.8) & 61.2 (64.8)
	\tabularnewline
	\texttt{  }$A_{package}^3$ & 65.8 (64.3) & 66 (64.6)
	\tabularnewline
	\texttt{  }$A_{package}^4$ & 52.5 (52) & 64.7 (58.7)
	\tabularnewline
	\texttt{  }$A_{package}^5$ & 52.5 (52) & 52.6 (58.7)
	\tabularnewline
	\texttt{  }$A_{package}^6$ & 50.4 (51.6) & 52.3 (52)
	\tabularnewline
	\texttt{  }$A_{type}^1$ & 51.4 (51.4) & 55.1 (53.7)
	\tabularnewline
	\texttt{  }$A_{type}^2$ & 54 (53.9) & 55.5 (54.3)
	\tabularnewline
	\texttt{  }$A_{type}^3$ & 56.8 (56.7) & 57 (56.9)
	\tabularnewline
	\texttt{  }$A_{type}^4$ & 57.1 (56.9) & 57.3 (57.1)
	\tabularnewline
	\texttt{  }$A_{type}^5$ & 57.4 (57.6) & 58 (57.9)
	\tabularnewline
	\texttt{  }$A_{type}^6$ & 57.2 (57.4) & 57.5 (57.5)
	\tabularnewline
	\texttt{  }Text-to-code Prob. & 55.7 & 55.8
	\tabularnewline
  	\bottomrule
\end{tabular}
\caption{Cross validation accuracy results for the coordinate term SVM
classifier (Code \& Corpus), as well as baselines using corpus distributional
similarity, string similarity, all corpus based features (Corpus Only), or all
code based features (Code Only), and all individual code based features. The
weighted version of the code based features (see Section~\ref{sec:coord-term-class}) is in parenthesis. Results are shown for
both the \emph{Coord} and
\emph{Coord-PMI} datasets.}
\label{tbl:classification-text-v-code}
\end{table}

Since the validity of the code based features above is directly related to the
success of the entity linking phase, each of the code based features
are used in the classifier once with the original value and a second time with the value
weighted by the text to code linking probability.

Of the noun pairs $\langle X,Y \rangle$ in our data, we keep only pairs for
which the linking probability $\hat{p}(X'|X)\cdot \hat{p}(Y'|Y)$ is greater than
$0.1$. Note that this guarantees that each noun must be mapped to at least one
class with non-zero probability. Next, we evaluate the string morphology and
its resemblance to a camel-case format, which is the acceptable formatting for
Java class names. We therefore select alphanumeric terms with at least two
upper-case and one lower-case characters. We name this set of noun pairs the
\emph{Coord} dataset.

A key assumption underlying statistical distributional similarity approaches is
that ``high-interest'' entities are associated with higher corpus frequencies,
therefore, given sufficient statistical evidence ``high-interest'' relations
can be extracted. In the software domain, real world factors may introduce biases in
a software-focused text corpus which may affect the corpus frequencies of
classes: e.g., users may discuss classes based on the clarity of their
API, the efficiency of their implementation, or simply if they
are fundamental in software introduced to novice users. Another
motivation for using grounded data, such as the class implementation, is that it
may highlight additional aspects of interest, for example, classes that
are commonly inherited from.
We therefore define a second noun dataset, \emph{Coord-PMI}, which
attempts to address this issue, in which noun
pairs are selected based on their pointwise mutual information (\emph{PMI}):
\begin{equation}
	\textrm{\emph{PMI}}(X,Y) = \log{\frac{p(X,Y)}{p(X)p(Y)}}
\end{equation}
where the frequency of the pair $\langle X,Y \rangle$ in the corpus is positive.
In this set we include coordinate term pairs with high \emph{PMI} scores, which
appear more rarely in the corpus and are therefore harder to predict using
standard NLP techniques. The negative set in this data are noun pairs which
appear frequently separately but do not appear as coordinate terms, and are
therefore marked by low \emph{PMI} scores.

To illustrate this point, we provide a sample of noun pairs with low and high
\emph{PMI} scores in Table~\ref{tbl:word-sample-pmi}, where pairs highlighted
with bold font are labeled as coordinate terms 
 in our data. We can see that the high
\emph{PMI} set contains pairs that are specific and interesting in the software
domain while not necessarily being frequent words in the general domain.
For example, some pairs seem to represent variable names
(e.g., \emph{$\langle$yearsPlayed, totalEarned$\rangle$}), others likely refer to
method names (e.g., \emph{$\langle$removeListener, addListener$\rangle$}).
Some pairs refer to Java classes, such as \emph{$\langle$JTextField,
JComboBox$\rangle$} whose implementation can be found in the Java
code repository. We can also see examples of pairs such as
\emph{$\langle$PostInsertEventListener, PostUpdateEventListener$\rangle$} which
are likely to be user-defined classes with a relationship to the Java class
\verb|java.util.EventListener|.
In contrast, the low \emph{PMI} set contains more general software terms (e.g.,
\emph{code, design, server, threads}).

\begin{table*} [t]
\centering
\rowcolors{1}{white}{gray!13}
\scalebox{0.97}{
\begin{tabular}{ p{5.3cm} p{5.8cm} l }
	\toprule
	Code Dist. Sim & $A_{package}^3$ & $A_{type}^5$
	\tabularnewline 
	\cmidrule{1-1} \cmidrule(l){2-2} \cmidrule(l){3-3}
	\small{$\langle$FileOutputStream,OutputStream$\rangle$} & 
	\small{$\langle$KeyEvent,KeyListener$\rangle$} &
	\small{$\langle$JMenuItem,JMenu$\rangle$}
	\tabularnewline 
	\small{$\langle$AffineTransform,AffineTransformOp$\rangle$} & 
	\small{$\langle$StyleConstants,SimpleAttributeSet$\rangle$} &
	\small{$\langle$JMenuItems,JMenu$\rangle$}
	\tabularnewline 
	\small{$\langle$GZIPOutputStream,}\newline
	\small{DeflaterOutputStream$\rangle$} &
	\small{$\langle$BlockQueue,ThreadPoolExecutor$\rangle$} &
	\small{$\langle$JMenuItems,JMenus$\rangle$} 
	\tabularnewline 
	\small{$\langle$OutputStream,DataOutputStream$\rangle$} & 
	\small{$\langle$BufferedImage,WritableRaster$\rangle$} &
	\small{$\langle$JLabel,DefaultTreeCellRenderer$\rangle$}
	\tabularnewline 
	\small{$\langle$AtomicInteger,AtomicIntegerArray$\rangle$} &
	\small{$\langle$MouseListener,MouseWheelListener$\rangle$} &
	\small{$\langle$JToggleButton,JRadioButtonMenu$\rangle$}
	\tabularnewline 
	\small{$\langle$ResourceBundle,ListResourceBundle$\rangle$} &
	\small{$\langle$DocumentBuilderFactory,}\newline
	\small{DocumentBuilder$\rangle$} &
	\small{$\langle$JFrame,JDialogs$\rangle$}
	\tabularnewline 
	\small{$\langle$setIconImages,setIconImage$\rangle$} &
	\small{$\langle$ActionListeners,FocusListeners$\rangle$} &
	\small{$\langle$JTable,JTableHeader$\rangle$}
	\tabularnewline 
	\small{$\langle$ComboBoxModel,}\newline
	\small{DefaultComboBoxModel$\rangle$} &
	\small{$\langle$DataInputStream,DataOutputStream$\rangle$} &
	\small{$\langle$JTextArea,JEditorPane$\rangle$}
	\tabularnewline 
	\small{$\langle$JTextArea,TextArea$\rangle$} &
	\small{$\langle$greaterOrEqualThan,lesserOrEqualThan$\rangle$} &
	\small{$\langle$JTextPane,JEditorPane$\rangle$}
	\tabularnewline 
	\small{$\langle$ServerSocketChannel,SocketChannel$\rangle$} &
	\small{$\langle$CopyOnWriteArrayList,}\newline
	\small{ConcurrentLinkedQueue$\rangle$} &
	\small{$\langle$JTextArea,JTable$\rangle$}
	\tabularnewline
  	\bottomrule
\end{tabular} 
}
\caption{Top ten coordinate terms predicted by classifiers
using one of the following features: code distributional similarity, package
hierarchy ancestry ($A_{package}^3$), and type hierarchy ancestry
($A_{type}^5$). All of the displayed
predictions are \emph{true}.}
\label{tbl:pairs-top-predicted}
\end{table*}

\section{Results}

\subsection{Classification and Feature Analysis.}

In Table~\ref{tbl:classification-text-v-code} we report the cross validation
accuracy of the coordinate term classifier (\emph{Code \& Corpus}) as well as 
baseline classifiers using corpus distributional similarity (\emph{Corpus
Dist. Sim.}), string similarity (\emph{String Sim.}), all corpus features
(\emph{All Corpus}), or all code features (\emph{All Code}).
Note that using all code features is significantly more successful on this data
than any of the corpus baselines (corpus baselines' accuracy is between 57\%-65\% whereas code-based
accuracy is over 80\%). When using both data sources, performance is
improved even further (to over 85\% on the \emph{Coord} dataset and 88\% on \emph{Coord-PMI}).

We provide an additional feature analysis in
Table~\ref{tbl:classification-text-v-code}, and report the
cross validation accuracy of classifiers using each single code feature.
Interestingly, code distributional similarity (\emph{Code Dist. Sim.}) is the
strongest single feature, and it is a significantly better predictor than corpus distributional
similarity, achieving around 67\% v.s. around 58\% for both datasets. 

\subsection{Evaluation by Manual Labeling.}

\begin{figure}[t]
\centering
	\includegraphics[width=0.45\textwidth]{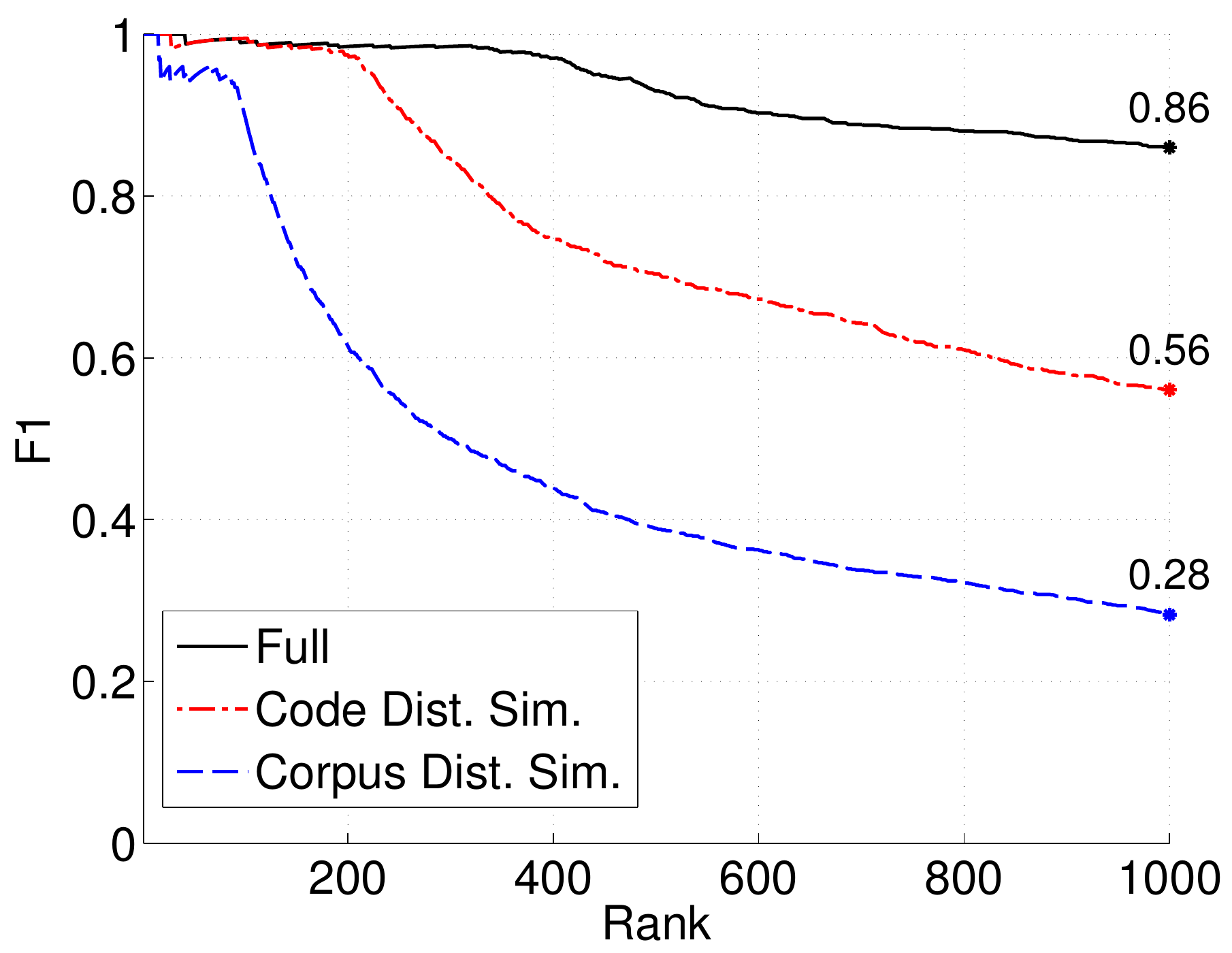}
	\caption{Manual Labeling Results. F1 results of the top 1000
	predicted coordinate terms by rank. The final data point in each line is
	labeled with the F1 score at rank 1000.}
  \label{fig:manualLabeling}
\end{figure}

The cross-validation results above are based on labels extracted using Hearst
conjunction patterns. In Figure~\ref{fig:manualLabeling} we provide an
additional analysis based on manual human labeling of samples from the
\emph{Coord-PMI} dataset, following a procedure similar to prior researchers
exploring semi-supervised methods for relation discovery
\cite{carlson2010toward,lao2011random}. After all development was complete,
we hand labeled the top 1000 coordinate term pairs according to the ranking by
our full classifier (using all code and corpus features) and the top 1000 pairs
predicted by the classifiers based on code and corpus distributional
similarities only. We report the F1 results of each classifier by
the rank of the predicted samples. According to our analysis, the F1 score for
the text and code distributional similarity classifiers degrades quickly after
the first 100 and 200 top ranked pairs, respectively. At rank 1000, the
score of the full classifier is at 86\%, whereas the code and text
classifiers are only at 56\% and 28\%.

To highlight the strength of each of the code based features, we provide in
Table~\ref{tbl:pairs-top-predicted} the top ten coordinate terms predicted using
the most successful code based features. For example, the top prediction using
type hierarchy ancestry ($A_{type}^5$) is
$\langle$JMenuItem, JMenu$\rangle$. Since \texttt{JMenu} extends
\texttt{JMenuItem}, the two classes indeed share many common interfaces and
classes. Alternatively, all of the top predictions using the package
hierarchy ancestry ($A_{package}^3$) are labels that have been matched
to pairs of classes that share at least 3 higher up package levels. So for
example, \texttt{BlockQueue} has been matched to
\texttt{java.util.concurrent.BlockingQueue} which was predicted as a coordinate
term of \texttt{ThreadPoolExecutor} which belongs in the same package. Using
code distributional similarity, one of the top predictions is
the pair $\langle$GZIPOutputStream, DeflaterOutputStream$\rangle$, which share
many common API methods such as \texttt{write}, \texttt{flush}, and
\texttt{close}. Many of the other top predicted pairs by this feature have been
mapped to the same class and therefore have the exact same context distribution.

\subsection{Taxonomy Construction.}
We visualize the coordinate term pairs predicted using our method (with all
features), by aggregating them into a graph where entities are nodes and edges
are determined by a coordinate term relation (Figure~\ref{fig:topology}). 
Graph edges are colored using the Louvain method \cite{louvain2008} for
community detection and an entity label's size is determined by its betweenness centrality
degree. We can see that high-level communities in this graph correspond
to class functionality, indicating that our method can be used to create an interesting
code taxonomy. 

Note that our predictions also highlight connections within functional groups
that cannot be found using the package or type taxonomies directly.
One example can be highlighted within the GUI functionality group.
\texttt{Listener} classes facilitate a response mechanism to GUI
\texttt{Actions}, such as \emph{pressing a button}, or \emph{entering text},
however, these classes belong in different packages than basic GUI components
for historical reasons.
In our graph, Action and Listener classes belong to the same communities of the
GUI components they are normally used with.

\section{Conclusions}

We have presented an approach for grounded discovery of coordinate term
relationships between text entities representing Java classes.
Using a simple entity linking method we map text entities to an underlying class
type implementation from the Java standard libraries. With this code-based
grounding, we extract information on the usage pattern of the class and its
location in the Java class and namespace hierarchies.
Our experimental evaluation shows that using only corpus distributional
similarity for the coordinate term prediction task is unsuccessful, achieving
prediction accuracy of around 58\%. However, adding information based on the
entities' software implementation improves accuracy dramatically to 88\%. Our
classifier has an F1 score of 86\% according to human labeling over the top 1000
predicted pairs. We have shown that our predictions can be used to build an
interesting code taxonomy which draws from the functional connections, common
usage patterns, and implementation details that are shared between classes.

\bibliography{coordinateTerm_arxiv2015}{}
\bibliographystyle{plain}
\end{document}